\documentclass[twoside,11pt]{article}

\usepackage{jmlr2eAdjusted}
\usepackage{amsmath}
\usepackage{dsfont}
\usepackage[]{algorithm2e}
\usepackage{graphicx}
\usepackage{booktabs}
\usepackage{enumerate}
\usepackage{siunitx}

\newcommand{\term}[1]{{\emph{#1}}} 
\newcommand{\ve}[1]{{\boldsymbol{#1}}} 
\newcommand{\ex}{\mathds{E}} 
\newcommand{\N}{\mathcal{N}} 
\newcommand{\ord}{\mathcal{O}} 
\newcommand{\diag}{\mbox{diag}} 
\newcommand{\Prb}{\mathbb{P}} 

\jmlrheading{?}{2017}{??--??}{05/17}{??/??}{Hildo Bijl, Thomas B. Sch\"on, Jan-Willem van Wingerden, Michel Verhaegen}

\ShortHeadings{A sequential Monte Carlo approach to Thompson sampling for Bayesian optimization}{Bijl}
\firstpageno{1}

\begin{document}
	
	\title{A sequential Monte Carlo approach to Thompson sampling for Bayesian optimization}
	
	\author{\name Hildo Bijl \email h.j.bijl@tudelft.nl \\
		\addr Delft Center for Systems and Control \\
		Delft University of Technology\\
		Delft, The Netherlands
		\AND
		\name Thomas B. Sch\"on \email thomas.schon@it.uu.se \\
		\addr Department of Information Technology \\
		Uppsala University \\
		Uppsala, Sweden
		\AND
		\name Jan-Willem van Wingerden \email j.w.vanwingerden@tudelft.nl \\
		\addr Delft Center for Systems and Control \\
		Delft University of Technology\\
		Delft, The Netherlands
		\AND
		\name Michel Verhaegen \email m.verhaegen@tudelft.nl \\
		\addr Delft Center for Systems and Control \\
		Delft University of Technology\\
		Delft, The Netherlands
	}
	
	\editor{??}
	
	\maketitle
	
	\begin{abstract}
		Bayesian optimization through Gaussian process regression is an effective method of optimizing an unknown function for which every measurement is expensive. It approximates the objective function and then recommends a new measurement point to try out. This recommendation is usually selected by optimizing a given acquisition function. After a sufficient number of measurements, a recommendation about the maximum is made. However, a key realization is that the maximum of a Gaussian process is not a deterministic point, but a random variable with a distribution of its own. This distribution cannot be calculated analytically. Our main contribution is an algorithm, inspired by sequential Monte Carlo samplers, that approximates this maximum distribution. Subsequently, by taking samples from this distribution, we enable Thompson sampling to be applied to (armed-bandit) optimization problems with a continuous input space. All this is done without requiring the optimization of a nonlinear acquisition function. Experiments have shown that the resulting optimization method has a competitive performance at keeping the cumulative regret limited.
	\end{abstract}
	
	
\section{Introduction}

Consider the problem of maximizing a continuous nonlinear reward function $f(\ve{x})$ (or equivalently minimizing a cost function) over a compact set $X_f$. In the case where $f(\ve{x})$ can be easily evaluated, where derivative data is available and where the function is convex (or concave), the solution is relatively straightforward, as is for instance discussed by~\citet{ConvexOptimization}. However, we will consider the case where convexity and derivative data are not known. In addition, every function evaluation is expensive and we can only obtain noisy measurements of the function. In this case, as was also indicated by~\cite{BlackBoxOptimization}, we need a data-driven approach to optimize the function.


The main idea is to try out certain inputs $\ve{x_1}, \ldots, \ve{x_n}$. After selecting a so-called \term{try-out input} $\ve{x_k}$, we feed it into the function and obtain a noisy measurement $y_k = f(\ve{x_k}) + \varepsilon$, with $\varepsilon = \N\left(0,\sigma_n^2\right)$. We then use all measurements obtained so far to make a Bayesian approximation of the function $f(\ve{x})$, based on which we choose the next try-out input $\ve{x_{k+1}}$. As such, this problem is known as \term{Bayesian optimization}~\citep{PracticleBayesianOptimization,BayesianOptimizationTutorial,BayesianOptimizationOverview}. In particular, we can approximate $f(\ve{x})$ through Gaussian process regression~\citep{GPBook}. This gives us a mean $\mu(\ve{x})$ and a standard deviation $\sigma(\ve{x})$ for our estimate of $f(\ve{x})$. The resulting optimization method is also known as \term{Gaussian process optimization}~\citep{OsborneGR:2009}. Bayesian methods like Gaussian process regression are known to efficiently deal with data, requiring only little data to make relatively accurate approximations. This makes these techniques suitable for a data-driven approach to problems in which data is expensive.

The main question is how to choose the try-out inputs $\ve{x_k}$. There are two different problem formulations available. In the first, after performing all $n$ measurements, we have to give a recommendation $\ve{\hat{x}^*}$ of what we believe is the true optimum $\ve{x^*}$. The difference $f^* - \hat{f}^*$ between the corresponding function values $f^* \equiv f(\ve{x^*})$ and $\hat{f}^* \equiv f(\ve{\hat{x}^*})$ is known as the \term{error} or the \term{instantaneous regret}. As such, this problem formulation is known as the \term{error minimization formulation} or also as the \term{probabilistic global optimization} problem. It is useful in applications like sensor placement~\citep{GPOThesis} and controller tuning in damage-free environments~\citep{GPOGaitOptimization,AutomaticLQRTuning}. These are all applications in which every try-out input (every experiment) has the same high set-up cost.

In the second problem formulation, our aim is to maximize the sum of all the rewards $f(\ve{x_1}), \ldots, f(\ve{x_n})$, which is known as the \term{value} $V$. Equivalently, we could also minimize the \term{(cumulative) regret}
\begin{equation}\label{eq:CumulativeRegret}
\sum_{k=1}^n \left(f^* - f(\ve{x_k})\right) = nf^* - V.
\end{equation}
This formulation is known as the \term{regret minimization formulation} or also as the \term{continuous armed bandit problem}. It is useful in applications like advertisement optimization~\citep{GPOAdvertisementHandling} or controller tuning for damage minimization (see Section~\ref{ss:ResultsMDGPOWindTurbine}). These are applications where the reward or cost of an experiment actually depends on the result of the experiment.  Because our applications fall in the latter category, we will focus on the regret minimization formulation in this paper. However, with the two formulations being similar, we also take error minimization strategies into account.


The main contribution of this paper is a new algorithm, inspired by the sequential Monte Carlo method~\citep{SMCPaper}, that approximates the maximum distribution. This algorithm can then be used to sample from the maximum distribution. This enables us to formulate an efficient Bayesian optimization algorithm with Thompson sampling for problems with a continuous input space, which is highly suitable for regret minimization problems. To the best of our knowledge, such an approach has not been applied before in literature and it is hence our main contribution.

We start by providing links to related work in Section~\ref{s:rw}. We will then present our main developments resulting in the Monte Carlo maximum distribution algorithm for approximating the distribution of the maximum in Section~\ref{s:MaximumDistribution}. We also analyze it and examine how we can use it to apply Thompson sampling. Experimental results are presented in Section~\ref{s:Results}, with conclusions and recommendations given in Section~\ref{s:Conclusions}.

\section{Related work}
\label{s:rw}

Both the error minimization and the regret minimization problems have been examined in literature before. In this section we examine the solutions that have already been proposed.

\subsection{Existing error minimization methods}

Several Bayesian optimization methods already exist. Good overviews are given by~\cite{PracticleBayesianOptimization,BayesianOptimizationTutorial,BayesianOptimizationOverview}, though we will provide a brief summary here. The recurring theme is that, when selecting the next input $\ve{x_k}$, we optimize some kind of \term{Acquisition Function}. In the literature, the discussion is mainly concerned with selecting and tuning an acquisition function.

The first to suggest the \term{Probability of Improvement} (PI) acquisition function was \cite{KushnerGPO}. This function is defined as $\mbox{PI}(\ve{x}) = \Prb(f(\ve{x}) \geq y_+)$, where $\Prb(A)$ denotes the probability of event $A$ to occur and $y_+$ denotes the highest value of the observation obtained so far. This was expanded by \cite{GlobalOptimizationBook,GOResponseSurfaces} to the form $\mbox{PI}(\ve{x}) = \Prb(f(\ve{x}) \geq y_+ + \xi)$, with $\xi$ being a tuning parameter trading off between exploration (high $\xi$) and exploitation (zero $\xi$).

Later on, \cite{BayesianMethodApplication} suggested an acquisition function which also takes the magnitude of the potential improvement into account. It is known as the \term{Expected Improvement} (EI) acquisition function $\mbox{EI}(\ve{x}) = \ex\left[\max(0,f(\ve{x}) - y_+)\right]$. Similar methods were used by others. For instance, multi-step lookahead was added by~\cite{GPOThesis}, a trust region to ensure small changes to the tried inputs $\ve{x_k}$ was used by~\cite{BayesianAscent}, and an additional exploration/exploitation parameter $\xi$ similar to the one used in the PI acquisition function was introduced by~\cite{BayesianOptimizationTutorial}. An analysis was performed by~\cite{EIAFAnalysis}.

Alternatively, \cite{GOStatisticalMethod} suggested the \term{Upper Confidence Bound} (UCB) acquisition function $\mbox{UCB}(\ve{x}) = \mu(\ve{x}) + \kappa \sigma(\ve{x})$. Here the parameter $\kappa$ determines the amount of exploration/exploitation, with high values resulting in more exploration. Often $\kappa = 2$ is used. The extreme case of $\kappa = 0$ is also known as the \term{Expected Value} (EV) acquisition function $\mbox{EV}(\ve{x}) = \mu(\ve{x})$. It applies only exploitation, so it is not very useful by itself. Methods to determine the value of $\kappa$ optimizing regret bounds were studied by~\cite{GPOInBanditSetting}.

A significant shift in focus was made through the introduction of the so-called \term{entropy search} method. This method was first developed by~\cite{GPOInformationalApproach}, although~\cite{EntropySearch} independently set up a similar method and introduced the name entropy search. The method was subsequently developed further as \term{predictive entropy search}\index{Predictive entropy search} by~\cite{PredictiveEntropySearch}. The main idea here is to look at the so-called \term{maximum distribution}: the probability $p_{\text{max}}(\ve{x}) \equiv \Prb(\ve{x} = \ve{x^*})$ that a certain point $\ve{x}$ equals the (unknown) optimum $\ve{x^*}$, or for continuous problems the corresponding probability density. We then focus on the relative entropy (the Kullback-Leibler divergence) of the maximum distribution compared to the flat probability density function over $X_f$. Initially this relative entropy is zero, but the more information we gain, the higher this relative entropy becomes. As such, we want to pick the try-out point $\ve{x_k}$ which is expected to increase the relative entropy the most.

At the same time, \term{portfolio methods} were developed with the aim to optimally use a whole assortment (a portfolio) of acquisition functions. These methods were introduced by~\cite{PortfolioMethods}, using results from~\cite{HedgeMethod,HedgingParameterFreeAlgorithm} and subsequently expanded on by~\cite{EntropySearchPortfolio}, who suggested to use the change in entropy as criterion to select recommendations.

\subsection{Existing regret minimization methods}

In the error minimization formulation, the focus is on obtaining as much information as possible. The regret minimization formulation is more involved, since it requires a trade-off between obtaining information and incurring costs (regret). Here, most of the research has focused on the case where the number of possible inputs $\ve{x}$ is finite. It is then known as the armed bandit problem and has been analyzed by for instance~\cite{BanditProblemAnalysis,BanditProblemRegretBounds,DeterministicBanditProblems}.

One of the more promising acquisition methods for the armed bandit problem is \term{Thompson sampling}. This method was first suggested by~\cite{ThompsonOriginalWork} and has more recently been analyzed by~\cite{ThompsonSamplingEvaluation,ThompsonSamplingAnalysis}. It is fundamentally different from other methods, because it does not use an acquisition function. Instead, we select an input point $\ve{x}$ as the next try-out point $\ve{x_k}$ with probability equal to the probability that $\ve{x}$ is the optimal input $\ve{x^*}$. This is equivalent to sampling $\ve{x_k}$ from the maximum distribution $p_{\text{max}}(\ve{x})$. Generally this distribution is not known though. When only finitely many different input points $\ve{x}$ are possible, the solution is to consider the vector $\ve{f} \equiv f(X)$ of all possible function outputs. Using Bayesian methods, we approximate $\ve{f}$ as a random variable, take a sample $\ve{\hat{f}}$ from it, find for which input point $\ve{x}$ this sample has its maximum, and subsequently use that input $\ve{x}$ as the next try-out point $\ve{x_k}$.

This method has proven to work well when the number of input points is finite. When there are infinitely many possible input points, like in continuous problems, it is impossible to sample from $\ve{f}$. This means that a new method to sample from the maximum distribution $p_{\text{max}}(\ve{x})$ is needed. However, in the existing literature this maximum distribution is not studied much at all. The idea of it was noted (but not evaluated) by~\cite{PracticleBayesianOptimization}. The maximum distribution was calculated by~\cite{GPOInformationalApproach} through a brute force method. An expansion to this was developed by~\cite{EntropySearch}, who used a method from~\cite{ExpectationPropagation} to approximate the minimum distribution. Though the approximation method used is quite accurate, it has a runtime of $\ord\left(n^4\right)$, making it infeasible to apply to most problems. An alternative method was described by~\cite{PredictiveEntropySearch} who approximated function samples of a Gaussian process through a finite number of basis functions and then optimized these function samples to generate samples from the maximum distribution. Though effective, this method requires solving a nonlinear optimization problem for each sample, which is computationally expensive and subject to the risk of finding only a local optimum.

\section{Finding the maximum distribution}\label{s:MaximumDistribution}

In this section we introduce an algorithm to find/approximate the distribution of the maximum of a Gaussian process. We then apply this algorithm to implement Thompson sampling.

\subsection{A Gaussian process and its maximum}\label{ss:GPAndItsMaximum}

Consider a function $f(\ve{x})$. We assume that we have taken $n$ measurements $y_i = f(\ve{x_i}) + \varepsilon$, where $\varepsilon \sim \N\left(0,\sigma_n^2\right)$ is Gaussian white noise. We merge all the measurement (training) input points $\ve{x_i}$ into a set $X$ and all the measured output values $y_i$ into a vector $\ve{y}$.

Now suppose that we want to predict the (noiseless) function values $\ve{f_*} = f(X_*)$ at a given set of trial (test) input points $X_*$. In this case we can use the standard GP regression equation from~\citet{GPBook}. We use a mean function $m(\ve{x})$ and a covariance function $k(\ve{x},\ve{x'})$, and we shorten $m(X_a)$ to $\ve{m_a}$ and $k(X_a,X_b)$ to $K_{ab}$ for any subscripts $a$ and $b$. (The subscript for the training set is omitted.) We then have
\begin{align}
\ve{f_*} & \sim \N\left(\ve{\mu_*}, \Sigma_{**}\right), \\
\ve{\mu_*} & = \ve{m_*} + K_{*}^T \left(K + \Sigma_n\right)^{-1} \left(\ve{y} - \ve{m}\right), \nonumber \\
\Sigma_{**} & = K_{**} - K_{*}^T \left(K + \Sigma_n\right)^{-1} K_{*}. \nonumber
\end{align}
A Gaussian process can be seen as a distribution over functions. That is, we can take samples of $\ve{f_*}$ and plot those as if they are functions, as is for instance done in Figure~\ref{fig:GPExampleWithSamples}. 
These sample functions generally have their maximum at different locations~$\ve{x^*}$. This implies that $\ve{x^*}$ is a random variable, and hence has a distribution $p_{\text{max}}(\ve{x})$. An example of this is shown in Figure~\ref{fig:GPExampleMaximumDistribution}.

The distribution $p_{\text{max}}(\ve{x})$ cannot be analytically calculated, but it can be approximated through various methods. The most obvious one is through brute force: for a finite number of trial input points $X_*$, we take a large number of samples $\ve{f_*}$, for each of these samples we find the location of the maximum, and through a histogram we determine the distribution of $\ve{x^*}$. This method is far from ideal as it is computationally very intensive, even for low-dimensional functions, but it is guaranteed to converge to the \term{true maximum distribution}.

For larger problems the brute force method is too computationally intensive, motivating the need for a way of approximating the maximum distribution. Methods to do so already exist, like those used by~\cite{EntropySearch,PredictiveEntropySearch}. However, these methods are all also computationally intensive for larger problems, and so a different way to approximate $p_{\text{max}}(\ve{x})$ would be beneficial.

\subsection{Approximating the maximum distribution}\label{ss:ApproximateMaximumDistribution}

We propose a new algorithm, inspired by Sequential Monte Carlo (SMC) samplers, to find the maximum distribution $p_{\text{max}}(\ve{x})$. Note that the algorithm presented here is \emph{not} an actual SMC sampler, but merely uses techniques also found in SMC samplers. For more background, see e.g.~\cite{SMCPaper,MCBook}.

The main idea is that we have $n_p$ so-called particles at positions $\ve{x^1}, \ldots, \ve{x^{n_p}}$. Each of these particles has a corresponding weight $w^1, \ldots, w^{n_p}$. Eventually these particles are supposed to converge to the maximum distribution, at which time we can approximate this distribution through kernel density estimation as
\begin{equation}\label{eq:KDEMaximumDistribution}
p_{\text{max}}(\ve{x}) \approx \frac{\sum_{i=1}^{n_p} w^i k_x(\ve{x},\ve{x^i})}{\sum_{i=1}^{n_p} w^i},
\end{equation}
with $k_x(\ve{x},\ve{x'})$ some manually chosen kernel. It is common
to make use of a squared exponential kernel with a small length scale.

Initially we distribute these particles $\ve{x^i}$ at random positions across the input space. That is, we sample the particles $\ve{x^i}$ from the flat distribution $q(\ve{x}) = c$. Note that, because we have assumed that the input space $X_f$ is compact, the constant $c$ is nonzero. 

To learn more about the position of the maximum, we will \term{challenge} existing particles. To challenge an existing particle $\ve{x^i}$, we first sample a number $n_c$ of random challenger particles $\ve{x_{c_1}^i}, \ldots, \ve{x_{c_{n_c}}^i}$ from a proposal distribution $q'(\ve{x})$. We then set up the joint distribution
\begin{equation}\label{eq:ChallengeDistribution}
\begin{bmatrix}
f(\ve{x^i}) \\
f(\ve{x_{c_1}^i}) \\
\vdots \\
f(\ve{x_{c_{n_c}}^i}) 
\end{bmatrix} \sim \N\left(\begin{bmatrix}
\mu(\ve{x^i}) \\
\mu(\ve{x_{c_1}^i}) \\
\vdots \\
\mu(\ve{x_{c_{n_c}}^i})
\end{bmatrix}, \begin{bmatrix}
\Sigma(\ve{x^i},\ve{x^i}) & \Sigma(\ve{x^i},\ve{x_{c_1}^i}) & \cdots & \Sigma(\ve{x^i},\ve{x_{c_{n_c}}^i}) \\
\Sigma(\ve{x_{c_1}^i},\ve{x^i}) & \Sigma(\ve{x_{c_1}^i},\ve{x_{c_1}^i}) & \cdots & \Sigma(\ve{x_{c_1}^i},\ve{x_{c_{n_c}}^i}) \\
\vdots & \vdots & \ddots & \vdots \\
\Sigma(\ve{x_{c_{n_c}}^i},\ve{x^i}) & \Sigma(\ve{x_{c_{n_c}}^i},\ve{x_{c_1}^i}) & \cdots & \Sigma(\ve{x_{c_{n_c}}^i},\ve{x_{c_{n_c}}^i})
\end{bmatrix}\right),
\end{equation}
and subsequently generate a sample $\begin{bmatrix}
\hat{f}^i & \hat{f}^i_{c_1} & \cdots & \hat{f}^i_{c_{n_c}}
\end{bmatrix}^T$ from it. Finally, we find the largest element from this vector. If this element equals $\hat{f}^i$, we do nothing. If, however, it equals $\hat{f}^i_{c_j}$, then we have $\hat{f}^i_{c_j} > \hat{f}^i$. In this case there is a challenger that has `beaten' the current particle and it takes its place. In other words, we replace the particle $\ve{x^i}$ by $\ve{x^i_{c_j}}$.

The challenger particle also has a weight associated with. In SMC methods this weight is usually given by
\begin{equation}
w_c^i = \frac{q(\ve{x_c^i})}{q'(\ve{x_c^i})}.
\end{equation}
However, to speed up convergence, we use a proposal distribution $q'(\ve{x})$ based on the ideas of mixture importance sampling and defensive importance sampling. Specifically, we use
\begin{equation}
q'(\ve{x}) = \alpha p_{\text{max}}(\ve{x}) + (1 - \alpha) q(\ve{x}).
\end{equation}
Here, $\alpha$ is manually chosen (often roughly $\frac{1}{2}$) and $p_{\text{max}}(\ve{x})$ is approximated through the mixture proposal distribution~\eqref{eq:KDEMaximumDistribution}, based on the current particle distribution. To generate a challenger particle $\ve{x_{c_j}^i}$, we hence randomly (according to the particle weights) select one of the particles $\ve{x^k}$. Then, in a part $\alpha$ of the cases, we sample $\ve{x_{c_j}^i}$ from $k_x(\ve{x},\ve{x^k})$, while in the remaining $(1 - \alpha)$ part of the cases, we sample $\ve{x_{c_j}^i}$ from $q(\ve{x})$. If we sample our challenger particles in this way, it is computationally more efficient to use the weight
\begin{equation}\label{eq:ChallengerWeight}
w_{c_j}^i = \frac{q(\ve{x_{c_j}^i})}{\alpha k_x(\ve{x_{c_j}^i},\ve{x^k}) + (1 - \alpha) q(\ve{x_{c_j}^i})}.
\end{equation}
Based on this formulation, we will challenge every existing particle once. This is called one \term{round} of challenges. Afterwards, we apply systematic resampling~\citep{Kitagawa:1996} to make sure all particles have the same weight again. We repeat this until the distribution of particles has mostly converged.

We call the resulting algorithm the \term{Monte Carlo Maximum Distribution} (MCMD) algorithm. Pseudo-code for it is given in Algorithm~\ref{alg:MCMD}.

\begin{algorithm}[htp]
	\label{alg:MCMD}
	\KwData{A known Gaussian process, user-defined parameters $n_p$, $n_c$, $\alpha$ and a kernel $k_x(\ve{x},\ve{x'})$.}
	\KwResult{An approximate distribution $p_{\text{max}}(\ve{x})$ of the optimal input $\ve{x^*}$, given through~\eqref{eq:KDEMaximumDistribution}.}
	\SetKwBlock{Initialize}{Initialization:}{end}
	\Initialize{
		\For{$i\leftarrow 1$ \KwTo $n_p$}{
			Sample $\ve{x^i}$ from $q(\ve{x})$. Assign $w^i = 1$.
		}
	}
	\SetKwBlock{Iterate}{Iteration:}{end}
	\Iterate{
		\Repeat{a sufficient number of rounds has passed}{
			Apply systematic resampling to all particles.\\
			\For{$i\leftarrow 1$ \KwTo $n_p$}{
				\For{$j\leftarrow 1$ \KwTo $n_c$}{
					Select a random particle $\ve{x^k}$, taking into account weights $w^k$.\\
					\eIf{we select a challenger based on $\ve{x^k}$ (probability $\alpha$)}{
						Sample a challenger particle $\ve{x_{c_j}^i}$ from the kernel $k_x(\ve{x},\ve{x^j})$.\\
					}{
					Sample a challenger particle $\ve{x_{c_j}^i}$ from the flat distribution $q(\ve{x})$.\\
				}
			}
			Sample a vector $\begin{bmatrix}
			\hat{f}^i & \hat{f}^i_{c_1} & \cdots & \hat{f}^i_{c_{n_c}}
			\end{bmatrix}^T$ based on~\eqref{eq:ChallengeDistribution} and find its maximum.\\
			\If{the maximum equals $\hat{f}^i_{c_j} > \hat{f}^i$}{
				Replace particle $\ve{x^i}$ by its challenger $\ve{x_{c_j}^i}$.\\
				Set the new weight $w^i$ according to~\eqref{eq:ChallengerWeight}.
			}
		}
	}
}
\caption{The Monte Carlo maximum distribution algorithm. Self-normalized importance sampling, mixture importance sampling and systematic resampling are used.}
\end{algorithm}

\subsection{Analysing the limit distribution of the algorithm}\label{ss:LimitDistributionAnalysis}

The distribution of the particles converges to a \term{limit distribution}. But does this limit distribution equal the true maximum distribution? We can answer this question for a few special cases.

First consider the case where $n_c \rightarrow \infty$. In this case, the algorithm is equivalent to the brute force method of finding the maximum distribution. Assuming that a sufficient number of particles $n_p$ is used, it hence is guaranteed to find the true maximum distribution directly, in only a single round of challenges.

Using $n_c \rightarrow \infty$ challenger particles is infeasible, because generating a sample from~\eqref{eq:ChallengeDistribution} takes $\ord(n_c^3)$ time. Instead, we consider a very simplified case with $n_c = 1$ and $\alpha = 0$. Additionally, we consider the discrete case, where there are finitely many possible input points $\ve{x_1}, \ldots, \ve{x_n}$. With finitely many points, we can use the kernel $k_x(\ve{x},\ve{x'}) = \delta(\ve{x} - \ve{x'})$, with $\delta(\ldots)$ the delta function. In this simplified case, we can analytically calculate the distribution that the algorithm converges to.

Consider the given Gaussian process. Let us denote the probability $\Prb(f(\ve{x_i}) > f(\ve{x_j}))$, based on the data in this Gaussian process, as $P_{ij}$. Here we have 
\begin{equation}
P_{ij} = \Prb(f(\ve{x_i}) > f(\ve{x_j})) = \Phi\left(\frac{\mu(\ve{x_i}) - \mu(\ve{x_j})}{\sqrt{\Sigma(\ve{x_i},\ve{x_i}) + \Sigma(\ve{x_j},\ve{x_j}) - 2\Sigma(\ve{x_i},\ve{x_j})}}\right),
\end{equation}
where $\Phi(\ldots)$ is the standard Gaussian cumulative density function. Through this expression we find the matrix $P$ element-wise. Additionally, we write the part of the particles that will eventually be connected to the input $\ve{x_i}$ as $p_i$. In this case, the resulting vector $\ve{p}$ (with elements $p_i$) can be shown to satisfy
\begin{equation}
\left(P - \diag\left(1_n P\right)\right) \ve{p} = \ve{0},
\end{equation}
where $1_n$ is an $n \times n$ matrix filled with ones, and $\diag(\ldots)$ is the function which sets all non-diagonal elements to zero. If we also use the fact that the sum of all probabilities $\ve{1}^T \ve{p}$ equals $1$, we can find $\ve{p}$ for this discrete problem.

If the algorithm would converge to the true maximum distribution in this simplified case (with $n_c = 1$) then we must have $p_i = p_{\text{max}}(\ve{x_i})$. In other words, the vector $\ve{p}$ would then describe the maximum distribution. However, since we can calculate the values $p_i$ analytically, while it is known to be impossible to find $p_{\text{max}}(\ve{x_i})$ like this, we already know that this is not the case. $p_i$ must be different from $p_{\text{max}}(\ve{x_i})$, and the algorithm hence does \emph{not} converge to the maximum distribution when $n_c = 1$. However, the example from Figure~\ref{fig:GPExampleMaximumDistribution} on page~\pageref{fig:GPExampleMaximumDistribution} does show that the algorithm gives a fair approximation. The limit distribution of the algorithm is generally less peaked than the true maximum distribution, which means it contains less information about where the maximum is (lower relative entropy) but overall its predictions are accurate. Furthermore, the difference will decrease when the variance present within the Gaussian process decreases, or when we raise $n_c$.

\subsection{Applying the MCMD algorithm for Thompson sampling}\label{ss:ApplyingMCMDForThompsonSampling}

We can now use the MCMD algorithm to apply Thompson sampling in a Gaussian process optimization setting. To do so, we sample an input point $\ve{x}$ from the approximated maximum distribution $p_{\text{max}}(\ve{x})$ whenever we need to perform a new measurement.

The downside of this method is that samples are not drawn from the true maximum distribution, but only from an approximation of it. However, the upside is that this approximation can be obtained by making simple comparisons between function values. No large matrix equations need to be solved or nonlinear function optimizations need to be performed, providing a significant computational advantage over other methods that approximate the maximum distribution.

\section{Experimental results}\label{s:Results}

Here we show the results of the presented algorithm. First we study how the MCMD algorithm works for a fixed one-dimensional Gaussian process. Then we apply it through Thompson sampling to optimize the same function, expand to a two-dimensional problem and finally apply it to a real-life application. Code related to the experiments can be found through~\citet{PhDThesisSourceCode} (Chapter 6).

\subsection{Execution of the MCMD algorithm}\label{ss:OneDimensionalProblem}

\begin{figure}[!th]
	\centering
	\includegraphics[width=.8\linewidth]{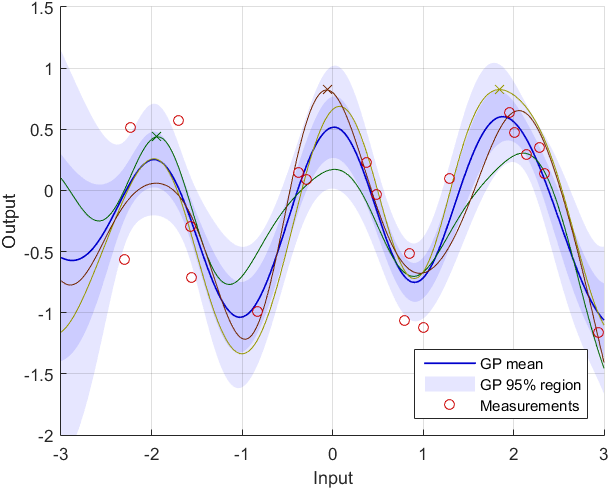}
	\caption{An example Gaussian process. The circles denote the measurements from which the GP was generated. The thick line denotes the (posterior) mean of the GP and the grey area represents the 95\% certainty region. The three thinner lines are samples from the GP distribution. It is worthwhile to note that they have their maximum values (the crosses) at very different positions.}
	\label{fig:GPExampleWithSamples}
\end{figure}

Consider the function
\begin{equation}
f(x) = \cos(3x) - \frac{1}{9}x^2 + \frac{1}{6}x. \label{eq:ExampleFunction}
\end{equation}
From this function, we take $20$ noisy measurements, at random locations in the interval $[-3,3]$, with $\sigma_n = 0.3$ as standard deviation of the white noise. We then apply GP regression with a squared exponential covariance function with predetermined hyperparameters. The subsequent GP approximating these measurements is shown in Figure~\ref{fig:GPExampleWithSamples}.

\begin{figure}[!th]
	\centering
	\includegraphics[width=.8\linewidth]{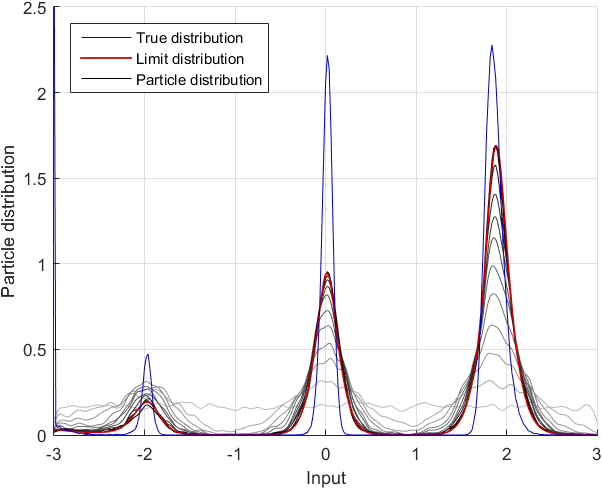}
	\caption{The maximum distribution for the Gaussian process shown in Figure~\ref{fig:GPExampleWithSamples}. The black/grey lines represent the approximate maximum distribution after $1, 2, \ldots, 10$ rounds of challenges for $n_p = 10 \thinspace 000$ particles. The red line is the limit distribution of the particles as derived in Section~\ref{ss:LimitDistributionAnalysis}. The blue line is the true maximum distribution, found through brute force methods.}
	\label{fig:GPExampleMaximumDistribution}
\end{figure}

We can apply the MCMD algorithm to approximate the maximum distribution $p_{\text{max}}(\ve{x})$ of this Gaussian process. This approximation, during successive challenge rounds of the algorithm with $\alpha = \frac{1}{2}$ and $n_c = 1$, is shown in Figure~\ref{fig:GPExampleMaximumDistribution}. (We always use $n_c = 1$ in these experiments, because it allows us to analytically calculate the limit distribution. For real-life experiments we would recommend larger values.) In this figure we see that the algorithm has mostly converged to the limit distribution after $n_r = 10$ rounds of challenges, but this limit distribution has a slightly higher entropy compared to the true maximum distribution. 

\subsection{Application to an optimization problem}\label{ss:Results1D}

We will now apply the newly developed method for Thompson sampling to Bayesian optimization. We will compare it with the UCB, the PI and the EI acquisition functions. After some tuning their parameters were set to $\kappa = 2$ and $\xi = 0.1$, which gave the best results we could obtain for these algorithms. To optimize these acquisition functions, we use a multi-start optimization method, because otherwise we occasionally end up with a local optimum of the acquisition function, resulting in a detrimental performance. We do not compare our results with entropy search or portfolio methods, because they are designed for the error minimization formulation.

The first problem we apply these methods to is the maximization of the function $f(x)$ of~\eqref{eq:ExampleFunction}. We use $n = 50$ input points $x_1, \ldots, x_n$ and look at the obtained regret. To keep the memory and runtime requirements of the GP regression algorithm somewhat limited, given the large number of experiments that will be run, we will apply the FITC approximation described by~\cite{SparseGPUnifyingView}, implemented in an online fashion according to~\cite{OnlineSparseGP}. As inducing input points, we use the chosen input points, but only when they are not within a distance $d_u$ (decreasing from $0.3$ to $0.02$ during the execution of the algorithm) of any already existing inducing input point. For simplicity the hyperparameters are assumed known and are hence fixed to reasonable values. Naturally, it is also possible to learn hyperparameters on-the-go as well, using the techniques described by~\citet{GPBook}.

\begin{figure}[!th]
	\centering
	\includegraphics[width=.8\linewidth]{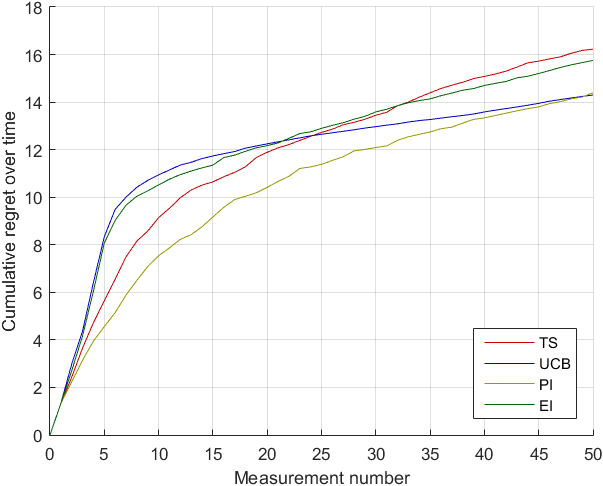}
	\caption{The cumulative regret~\eqref{eq:CumulativeRegret} of
          the various Bayesian optimization algorithms for the
          function in~\eqref{eq:ExampleFunction}. Results shown are the mean performance of fifty complete runs of each algorithm.}
	\label{fig:CumulativeRegret1D}
\end{figure}

The result is shown in Figure~\ref{fig:CumulativeRegret1D}. In this particular case, it seems that Thompson sampling and the PI acquisition function applied mostly exploitation: they have a better short term performance. On the other hand, the UCB and EI acquisition functions apply more exploration: the cost of quickly exploring is higher, but because the optimum is found sooner, it can also be exploited sooner.

It should also be noted that all algorithms occasionally end up in the wrong optimum (near $x = 2$). This can be seen from the fact that the regret graph does not level out. For this particular problem, the UCB acquisition function seems to be the best at avoiding the local optima, but it still falls for them every now and then. As noted earlier, only Thompson sampling has the guarantee to escape local optima given infinitely many measurements.

\subsection{Extension to a two-dimensional problem}\label{ss:ResultsMDGPO2D}

\begin{figure}[!th]
	\centering
	\includegraphics[width=.8\linewidth]{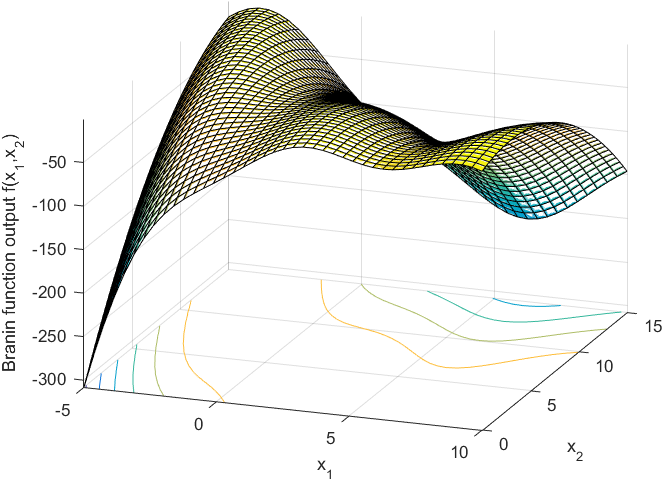}
	\caption{The (negative) Branin function, defined by~\eqref{eq:BraninFunction}.}
	\label{fig:BraninFunction}
\end{figure}

Next, we apply the optimization methods to a two-dimensional problem. We will minimize the well-known Branin function from (among others)~\cite{BraninFunction}. Or equivalently, we maximize the negative Branin function,
\begin{align}
f(x_1,x_2) & = -\left(x_2 - \frac{51 x_1^2}{40\pi^2} + \frac{5x_1}{\pi} - 6\right)^2 - 10 \left(1 - \frac{1}{8\pi}\right) \cos(x_1) - 10, \label{eq:BraninFunction}
\end{align}
where $x_1 \in [-5,10]$ and $x_2 \in [0,15]$. This function is shown in Figure~\ref{fig:BraninFunction}. We can find analytically that the optima occur at $\left(-\pi,\frac{491}{40}\right)$, $\left(\pi,\frac{91}{40}\right)$ and $\left(3\pi, \frac{99}{40}\right)$, all with value $-\frac{5}{4\pi}$.

\begin{figure}[!th]
	\centering
	\includegraphics[width=.8\linewidth]{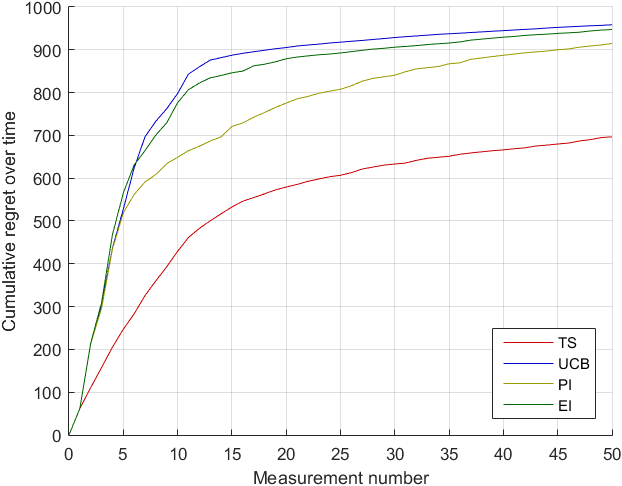}
	\caption{The cumulative regret~\eqref{eq:CumulativeRegret} of the various Bayesian optimization algorithms for the Branin function~\eqref{eq:BraninFunction}. Results shown are the mean performance of fifty complete runs of each algorithm.}
	\label{fig:CumulativeRegret2D}
\end{figure}

The performance of the various optimization methods, averaged out over fifty full runs, is shown in Figure~\ref{fig:CumulativeRegret2D}. Here we see that Thompson sampling now performs significantly better at keeping the regret small compared to the UCB ($\kappa = 2$), the PI and the EI ($\xi = 2$) acquisition functions. We can find the reason behind this, if we look at which try-out points the various algorithms select. When we do (not shown here), we see that all acquisition functions often try out points at the border of the input space, while Thompson sampling does not. In particular, the acquisition functions (nearly) always try out all four corners of the input space, including the very detrimental point $\left(-5,0\right)$. It is this habit which makes these acquisition functions perform worse on this specific problem.

\begin{figure}[!th]
	\centering
	\includegraphics[width=.8\linewidth]{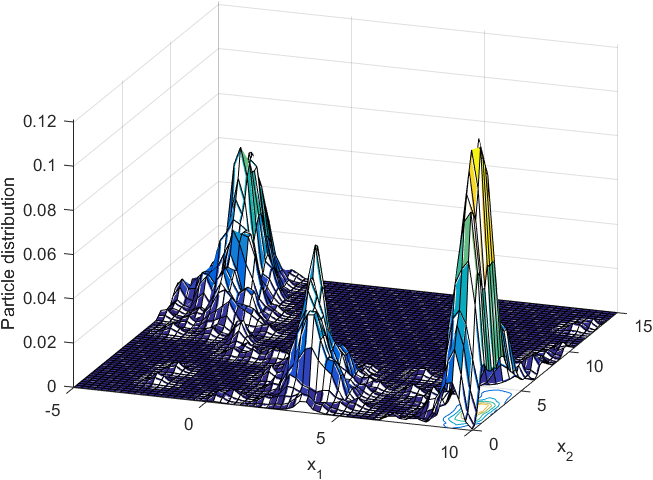}
	\caption{The probability distribution of the maximum of the GP approximating the Branin function, after generating measurements according to Thompson sampling. The three optimums have been identified, some stray particles still reside in the lesser explored regions, and no particles remain in the part of the input space that has been explored but was found suboptimal.}
	\label{fig:MaximumDistribution2D}
\end{figure}

Other than this, it is also interesting to note that in all examined runs, all optimization methods find either two or three of the optima. So while multiple optima are always found, it does regularly happen that one of the three optima is not found. All of the methods have shown to be susceptible to this. In addition, the three acquisition functions have a slightly lower average recommendation error than Thompson sampling, but since all optimization methods find various optimums, the difference is negligible. On the flip side, an advantage of using the MCMD algorithm is that it can provide us with the posterior distribution of the maximum, given all the measurement data. An example of this is shown in Figure~\ref{fig:MaximumDistribution2D}.

\subsection{Optimizing a wind turbine controller}\label{ss:ResultsMDGPOWindTurbine}

Finally we test our algorithm on an application: data-based controller tuning for load mitigation within a wind turbine. More specifically, we use a linearized version of the so-called TURBU model, described by~\cite{TURBU}. TURBU is a fully integrated wind turbine design and analysis tool. It deals with aerodynamics, hydrodynamics, structural dynamics and control of modern three bladed wind turbines, and as such gives very similar results as an actual real-life wind turbine.

We will consider the case where trailing edge flaps have been added to the turbine blades. These flaps should then be used to reduce the vibration loads within the blades. To do so, the Root Bending Moment (RBM) of the blades is used as input to the control system.

To determine the effectiveness of the controller, we look at two quantities. The first is the Damage Equivalent Load (DEL; see~\cite{DamageEquivalentLoad}). The idea here is that the blades are subject to lots of vibrations, some with large magnitudes and some with small magnitudes. For fatigue damage, large oscillations are much more significant. To take this into account, we look at which \SI{1}{Hz} sinusoidal load would result in the same fatigue damage as all measured oscillations put together. To accomplish this, the RBM signal is separated into individual oscillations using the rainflow algorithm~\citep{RainflowApplication}. We then use Miner's rule~\citep{RandomVibrations}, applying a W\"ohler exponent of $m = 11$ for the glass fiber composite blades~\citep{WindTurbineFrequencyAnalysis}, to come up with an equivalent \SI{1}{Hz} load.

The second quantity to optimize is the mean rate of change of the input signal. The reason here is that the lifetime of bearings is often expressed in the number of revolutions, or equivalently in the angular distance traveled, and dividing this distance traveled by the time passed will result in the mean rate of change of the flap angle. The eventual performance score for a controller will now be a linearly weighted sum of these two parameters, where a lower score is evidently better. 

As controller, we apply a proportional controller. That is, we take the RBM in the fixed reference frame (so after applying a Coleman transformation; see~\cite{ColemanTransformation}) and feed the resulting signal, multiplied by a constant gain, to the blade flaps. Since the wind turbine has three blades, there are three gains which we can apply. The first of these, the collective flap mode, interferes with the power control of the turbine. We will hence ignore this mode and only tune the gains of the tilt and yaw modes. Very low gains (in the order of $10^{-8}$) will result in an inactive controller which does not reduce the RBM, while very high gains (in the order of $10^{-5}$) will react to every small bit of turbulence, resulting in an overly aggressive controller with a highly varying input signal. Both are suboptimal, and the optimal controller will have gains somewhere between these two extreme values.

\begin{figure}[!th]
	\centering
	\includegraphics[width=.8\linewidth]{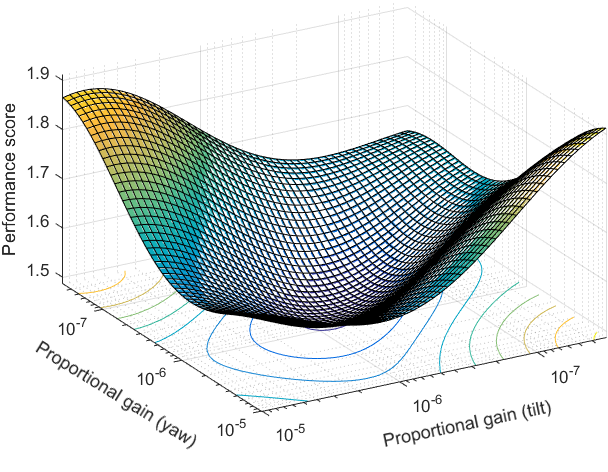}
	\caption{An approximation of the wind turbine controller score with respect to the controller gain. This approximation was made by taking $500$ random points and applying a GP regression algorithm to the resulting measurements.}
	\label{fig:Turbine2DCostFunction}
\end{figure}

To learn more about the actual score function, we can apply a brute force method -- just applying 500 random controller settings -- and apply GP regression. This gives us Figure~\ref{fig:Turbine2DCostFunction}. Naturally, this is not possible in real life as it would cause unnecessary damage to the wind turbine. It does tell us, however, that the score function is mostly convex and that there does not seem to exist any local optimums.

\begin{figure}[!th]
	\centering
	\includegraphics[width=.8\linewidth]{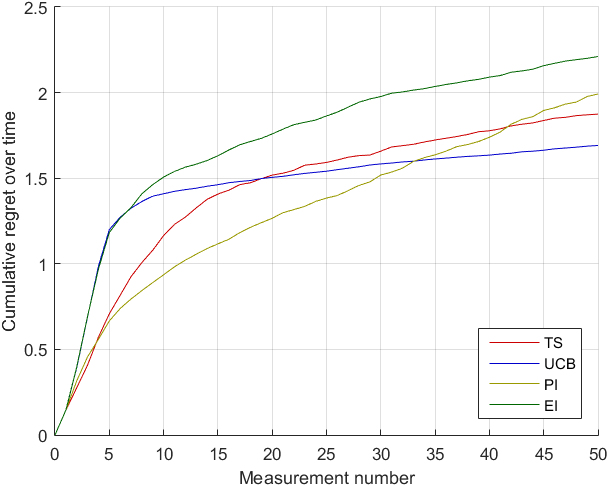}
	\caption{The cumulative regret~\eqref{eq:CumulativeRegret} of the various Bayesian optimization algorithms for the wind turbine controller. Results shown are the mean performance of fifty complete runs of each algorithm.}
	\label{fig:Turbine2DResults}
\end{figure}

The results from the Bayesian optimization experiments, which are remarkably similar to earlier experiments, are shown in Figure~\ref{fig:Turbine2DResults}. (We used $\kappa = 1$ and $\xi = 0.005$ here.) They once more show that Thompson sampling has a competitive performance at keeping the regret limited. A similar experiment, though with far fewer measurements, has been performed on a scaled wind turbine placed in a wind tunnel, and the results there were similar as well. See~\cite{PhDThesis} for further details on this wind tunnel test.

\section{Conclusions and recommendations}\label{s:Conclusions}

We have introduced the MCMD algorithm, which uses particles to approximate the distribution of the maximum of a Gaussian process. This particle approximation can then be used to set up a Bayesian optimization method using Thompson sampling. Such optimization methods are suitable for tuning the parameters of systems with large amounts of uncertainty in an online data-based way. As an example, we have tuned the controller gains of a wind turbine simulation to reduce the fatigue load using performance data that was obtained during the operation of the wind turbine.

The main advantage of Thompson sampling with the MCMD algorithm is that it does not require the optimization of a nonlinear function to select the next try-out point. In addition, it has shown to have a competitive performance at keeping the cumulative regret limited. However, we cannot conclude that Thompson sampling, or any other optimization method, works better than its competitors. Which method works well depends on a variety of factors, like how much the method has been fine-tuned to the specific function that is being optimized, as well as which function is being optimized in the first place. Also the number of try-out points used matters, where a lower number gives the advantage to exploitation-based methods, while a higher number benefits the more exploration-based methods. It is for this very reason that any claim that a Bayesian optimization works better than its competitors may be accepted only after very careful scrutiny.

\acks{This research is supported by the Dutch Technology Foundation STW, which is part of the Netherlands Organisation for Scientific Research (NWO), and which is partly funded by the Ministry of Economic Affairs. The work was also supported by the Swedish research Council (VR) via the projects \emph{NewLEADS - New Directions in Learning Dynamical Systems} and \emph{Probabilistic modeling of dynamical systems} (Contract number: 621-2016-06079, 621-2013-5524) and by the Swedish Foundation for Strategic Research (SSF) via the project \emph{ASSEMBLE} (Contract number: RIT15-0012).}

\vskip 0.2in
\bibliography{bibliography}

\begin{thebibliography}{47}
\providecommand{\natexlab}[1]{#1}
\providecommand{\url}[1]{\texttt{#1}}
\expandafter\ifx\csname urlstyle\endcsname\relax
  \providecommand{\doi}[1]{doi: #1}\else
  \providecommand{\doi}{doi: \begingroup \urlstyle{rm}\Url}\fi

\bibitem[Agrawal and Goyal(2012)]{ThompsonSamplingAnalysis}
Shipra Agrawal and Navin Goyal.
\newblock Analysis of {Thompson} sampling for the multi-armed bandit problem.
\newblock In \emph{JMLR Workshop and Conference Proceedings}, volume~23, pages
  39.1--39.26, 2012.

\bibitem[Auer et~al.(1995)Auer, Cesa-Bianchi, Freund, and
  Schapire]{HedgeMethod}
Peter Auer, Nicolo Cesa-Bianchi, Yoav Freund, and Robert~E. Schapire.
\newblock Gambling in a rigged casino: {The} adversarial multi-armed bandit
  problem.
\newblock In \emph{Proceedings of the 36th Annual Symposium on Foundations of
  Computer Science}, pages 322--331, 1995.

\bibitem[Bijl(2017{\natexlab{a}})]{PhDThesis}
Hildo Bijl.
\newblock \emph{Gaussian process regression techniques}.
\newblock PhD thesis, Delft University of Technology, 2017{\natexlab{a}}.

\bibitem[Bijl(2017{\natexlab{b}})]{PhDThesisSourceCode}
Hildo Bijl.
\newblock Gaussian progress regression techniques source code,
  2017{\natexlab{b}}.
\newblock URL \url{https://github.com/HildoBijl/GPRT}.

\bibitem[Bijl et~al.(2015)Bijl, {van Wingerden}, Sch\"on, and
  Verhaegen]{OnlineSparseGP}
Hildo Bijl, Jan-Willem {van Wingerden}, Thomas~B. Sch\"on, and Michel
  Verhaegen.
\newblock Online sparse {G}aussian process regression using {FITC} and {PITC}
  approximations.
\newblock In \emph{Proceedings of the IFAC symposium on System Identification,
  SYSID, Beijing, China}, October 2015.

\bibitem[Boyd and Vandenberghe(2004)]{ConvexOptimization}
Stephen Boyd and Lieven Vandenberghe.
\newblock \emph{Convex Optimization}.
\newblock Cambridge University Press, 2004.

\bibitem[Brochu et~al.(2010)Brochu, Cora, and {de
  Freitas}]{BayesianOptimizationTutorial}
Eric Brochu, Vlad~M Cora, and Nando {de Freitas}.
\newblock A tutorial on {B}ayesian optimization of expensive cost functions,
  with application to active user modeling and hierarchical reinforcement
  learning.
\newblock Technical report, University of British Columbia, 2010.

\bibitem[Candela and Rasmussen(2005)]{SparseGPUnifyingView}
Joaquin~Q. Candela and Carl~E. Rasmussen.
\newblock A unifying view of sparse approximate {G}aussian process regression.
\newblock \emph{Journal of Machine Learning Research}, 6:\penalty0 1939--1959,
  2005.

\bibitem[Chapelle and Li(2011)]{ThompsonSamplingEvaluation}
Olivier Chapelle and Lihong Li.
\newblock An empirical evaluation of {Thompson} sampling.
\newblock In \emph{Advances in Neural Information Processing Systems},
  volume~24, pages 2249--2257. Curran Associates, Inc., 2011.

\bibitem[Chaudhuri et~al.(2009)Chaudhuri, Freund, and
  Hsu]{HedgingParameterFreeAlgorithm}
Kamalika Chaudhuri, Yoav Freund, and Daniel~J. Hsu.
\newblock A parameter-free hedging algorithm.
\newblock In \emph{Advances in Neural Information Processing Systems},
  volume~22, pages 297--305, 2009.

\bibitem[Cox and John(1997)]{GOStatisticalMethod}
Dennis~D. Cox and Susan John.
\newblock {SDO}: A statistical method for global optimization.
\newblock In \emph{Multidisciplinary Design Optimization: State-of-the-Art},
  pages 315--329, 1997.

\bibitem[de~Freitas et~al.(2012)de~Freitas, Smola, and
  Zoghi]{DeterministicBanditProblems}
Nando de~Freitas, Alex Smola, and Masrour Zoghi.
\newblock Regret bounds for deterministic gaussian process bandits.
\newblock Technical report, arXiv.org, 2012.

\bibitem[Del~Moral et~al.(2006)Del~Moral, Doucet, and Jasra]{SMCPaper}
Pierre Del~Moral, Arnaud Doucet, and Ajay Jasra.
\newblock Sequential {M}onte {C}arlo samplers.
\newblock \emph{Journal of the Royal Statistical Society: Series B (Statistical
  Methodology)}, 68\penalty0 (3):\penalty0 411--436, 2006.

\bibitem[Dixon and Szeg\"o(1978)]{BraninFunction}
L.C.W. Dixon and G.P. Szeg\"o.
\newblock The global optimisation problem: an introduction.
\newblock In L.C.W. Dixon and G.P. Szeg\"o, editors, \emph{Towards global
  optimization}, volume~2, pages 1--15. North-Holland Publishing, 1978.

\bibitem[Freebury and Musial(2000)]{DamageEquivalentLoad}
Gregg Freebury and Walter Musial.
\newblock Determining equivalent damage loading for full-scale wind turbine
  blade fatigue tests.
\newblock In \emph{Proceedings of the 19th American Society of Mechanical
  Engineers (ASME) Wind Energy Symposium, Reno, Nevada}, 2000.

\bibitem[Gr\"unew\"alder et~al.(2010)Gr\"unew\"alder, Audibert, Opper, and
  Shawe-Taylor]{BanditProblemRegretBounds}
Steffen Gr\"unew\"alder, Jean-Yves Audibert, Manfred Opper, and John
  Shawe-Taylor.
\newblock Regret bounds for {Gaussian} process bandit problems.
\newblock In \emph{Proceedings of the Thirteenth International Conference on
  Artificial Intelligence and Statistics (AISTATS)}, 2010.

\bibitem[Hennig and Schuler(2012)]{EntropySearch}
Philipp Hennig and Christian~J. Schuler.
\newblock Entropy search for information-efficient global optimization.
\newblock \emph{Journal of Machine Learning Research}, 13:\penalty0 1809--1837,
  2012.

\bibitem[Hern\'andez-Lobato et~al.(2014)Hern\'andez-Lobato, Hoffman, and
  Ghahramani]{PredictiveEntropySearch}
Jos\'e~M. Hern\'andez-Lobato, Matthew~W. Hoffman, and Zoubin Ghahramani.
\newblock Predictive entropy search for efficient global optimization of
  black-box functions.
\newblock In \emph{Advances in Neural Information Processing Systems 27}.
  Curran Associates, Inc., 2014.

\bibitem[Hoffman et~al.(2011)Hoffman, Brochu, and de~Freitas]{PortfolioMethods}
Matthew Hoffman, Eric Brochu, and Nando de~Freitas.
\newblock Portfolio allocation for {Bayesian} optimization.
\newblock In \emph{Uncertainty in Artificial Intelligence (UAI)}, pages
  327--336, 2011.

\bibitem[Jones(2001)]{GOResponseSurfaces}
Donald~R. Jones.
\newblock A taxonomy of global optimization methods based on response surfaces.
\newblock \emph{Journal of Global Optimization}, 21\penalty0 (4):\penalty0
  345--383, 2001.

\bibitem[Jones et~al.(1998)Jones, Schonlau, and Welch]{BlackBoxOptimization}
Donald~R. Jones, Matthias Schonlau, and William~J. Welch.
\newblock Efficient global optimization of expensive black-box functions.
\newblock \emph{Journal of Global Optimization}, 13\penalty0 (4):\penalty0
  455--492, 1998.

\bibitem[Kitagawa(1996)]{Kitagawa:1996}
G.~Kitagawa.
\newblock Monte {C}arlo filter and smoother for non-{G}aussian nonlinear state
  space models.
\newblock \emph{Journal of {C}omputational and {G}raphical {S}tatistics},
  5\penalty0 (1):\penalty0 1--25, 1996.

\bibitem[Kleinberg(2004)]{BanditProblemAnalysis}
Robert~D. Kleinberg.
\newblock Nearly tight bounds for the continuum-armed bandit problem.
\newblock In \emph{Advances in Neural Information Processing Systems 17}, pages
  697--704. MIT Press, 2004.

\bibitem[Kushner(1964)]{KushnerGPO}
Harold~J. Kushner.
\newblock A new method of locating the maximum point of an arbitrary multipeak
  curve in the presence of noise.
\newblock \emph{Journal of Basic Engineering}, 86\penalty0 (1):\penalty0
  97--106, 1964.

\bibitem[Lizotte et~al.(2007)Lizotte, Wang, Bowling, and
  Schuurmans]{GPOGaitOptimization}
Daniel Lizotte, Tao Wang, Michael Bowling, and Dale Schuurmans.
\newblock Automatic gait optimization with {Gaussian} process regression.
\newblock In \emph{Proceedings of the 20th International Joint Conference on
  Artifical Intelligence}, pages 944--949, 2007.

\bibitem[Lizotte(2008)]{PracticleBayesianOptimization}
Daniel~James Lizotte.
\newblock \emph{Practical {B}ayesian Optimization}.
\newblock PhD thesis, University of Alberta, 2008.

\bibitem[Marco et~al.(2016)Marco, Hennig, Bohg, Schaal, and
  Trimpe]{AutomaticLQRTuning}
Alonso Marco, Philipp Hennig, Jeannette Bohg, Stefan Schaal, and Sebastian
  Trimpe.
\newblock Automatic {LQR} tuning based on {G}aussian process global
  optimization.
\newblock In \emph{Proceedings of the IEEE International Conference on Robotics
  and Automation (ICRA) 2016}. IEEE, May 2016.

\bibitem[Minka(2001)]{ExpectationPropagation}
Thomas~P. Minka.
\newblock Expectation propagation for approximate bayesian inference.
\newblock In \emph{Proceedings of the Seventeenth Conference on Uncertainty in
  Artificial Intelligence}, 2001.

\bibitem[Mockus et~al.(1978)Mockus, Tiesis, and
  Zilinskas]{BayesianMethodApplication}
Jonas Mockus, Vytautas Tiesis, and Antanas Zilinskas.
\newblock \emph{The application of {B}ayesian methods for seeking the
  extremum}.
\newblock Elsevier, Amsterdam, 1978.

\bibitem[Nies{\l}ony(2009)]{RainflowApplication}
Adam Nies{\l}ony.
\newblock Determination of fragments of multiaxial service loading strongly
  influencing the fatigue of machine components.
\newblock \emph{Mechanical Systems and Signal Processing}, 23\penalty0
  (8):\penalty0 2712--2721, November 2009.

\bibitem[Osborne et~al.(2009)Osborne, Garnett, and Roberts]{OsborneGR:2009}
M.~A. Osborne, R.~Garnett, and S.~J. Roberts.
\newblock Gaussian processes for global optimization.
\newblock In \emph{Proceedings of the 3rd international conference on learning
  and intelligent optimization ({LION}3)}, pages 1--15, Trento, Italy, January
  2009.

\bibitem[Osborne(2010)]{GPOThesis}
Michael Osborne.
\newblock \emph{{B}ayesian {G}aussian Processes for Sequential Prediction,
  Optimisation and Quadrature}.
\newblock PhD thesis, University of Oxford, 2010.

\bibitem[Owen(2013)]{MCBook}
Art~B. Owen.
\newblock {M}onte {C}arlo theory, methods and examples.
\newblock Unpublished manuscript, 2013.

\bibitem[Pandey and Olston(2007)]{GPOAdvertisementHandling}
Sandeep Pandey and Christopher Olston.
\newblock Handling advertisements of unknown quality in search advertising.
\newblock In \emph{Advances in Neural Information Processing Systems},
  volume~19, pages 1065--1072. MIT Press, 2007.

\bibitem[Park and Law(2015)]{BayesianAscent}
Jinkyoo Park and Kincho~H. Law.
\newblock {B}ayesian {A}scent ({BA}): A data-driven optimization scheme for
  real-time control with application to wind farm power maximization.
\newblock \emph{IEEE Transactions on Control Systems Technology}, November
  2015.

\bibitem[Rasmussen and Williams(2006)]{GPBook}
Carl~E. Rasmussen and Christopher~K.I. Williams.
\newblock \emph{{G}aussian Processes for Machine Learning}.
\newblock MIT Press, 2006.

\bibitem[Savenije and Peeringa(2009)]{WindTurbineFrequencyAnalysis}
Feike~J. Savenije and J.M. Peeringa.
\newblock Aero-elastic simulation of offshore wind turbines in the frequency
  domain.
\newblock Technical Report Report ECN-E-09-060, Energy research centre ECN, The
  Netherlands, 2009.

\bibitem[Shahriari et~al.(2014)Shahriari, Wang, Hoffman, Bouchard-C\^ot\'e, and
  de~Freitas]{EntropySearchPortfolio}
Bobak Shahriari, Ziyu Wang, Matthew~W. Hoffman, Alexandre Bouchard-C\^ot\'e,
  and Nando de~Freitas.
\newblock An entropy search portfolio for {Bayesian} optimization.
\newblock Technical report, University of Oxford, 2014.

\bibitem[Shahriari et~al.(2016)Shahriari, Swersky, Wang, Adams, and
  de~Freitas]{BayesianOptimizationOverview}
Bobak Shahriari, Kevin Swersky, Ziyu Wang, Ryan~P. Adams, and Nando de~Freitas.
\newblock Taking the human out of the loop: A review of {B}ayesian
  optimization.
\newblock \emph{Proceedings of the IEEE}, 104\penalty0 (1):\penalty0 148--175,
  January 2016.

\bibitem[Srinivas et~al.(2012)Srinivas, Krause, Kakade, and
  Seeger]{GPOInBanditSetting}
Niranjan Srinivas, Andreas Krause, Sham~M. Kakade, and Matthias~W. Seeger.
\newblock Information-theoretic regret bounds for {G}aussian process
  optimization in the bandit setting.
\newblock \emph{IEEE Transactions on Information Theory}, 58\penalty0
  (5):\penalty0 3250 -- 3265, May 2012.

\bibitem[Thompson(1933)]{ThompsonOriginalWork}
William~R. Thompson.
\newblock On the likelihood that one unknown probability exceeds another in
  view of the evidence of two samples.
\newblock \emph{Biometrika}, 25\penalty0 (3/4):\penalty0 285--294, 1933.

\bibitem[Torn and Zilinskas(1989)]{GlobalOptimizationBook}
Aimo Torn and Antanas Zilinskas.
\newblock \emph{Global Optimization}.
\newblock Springer-Verlag New York, Inc., 1989.

\bibitem[van {E}ngelen and Braam(2004)]{TURBU}
T.~van {E}ngelen and H.~Braam.
\newblock {TURBU} {O}ffshore; {C}omputer program for frequency domain analysis
  of horizontal axis offshore wind turbines - {I}mplementation.
\newblock Technical Report Report ECN-C-04-079, ECN, 2004.

\bibitem[{van}~Solingen and {van}~Wingerden(2015)]{ColemanTransformation}
E.~{van}~Solingen and J.~W. {van}~Wingerden.
\newblock Linear individual pitch control design for two-bladed wind turbines.
\newblock \emph{Wind Energy}, 18:\penalty0 677--697, 2015.

\bibitem[Vazquez and Bect(2010)]{EIAFAnalysis}
Emmanuel Vazquez and Julien Bect.
\newblock Convergence properties of the expected improvement algorithm with
  fixed mean and covariance functions.
\newblock \emph{Journal of Statistical Planning and Inference}, 140\penalty0
  (11):\penalty0 3088--3095, 2010.

\bibitem[Villemonteix et~al.(2009)Villemonteix, Vazquez, and
  Walter]{GPOInformationalApproach}
Julien Villemonteix, Emmanuel Vazquez, and Eric Walter.
\newblock An informational approach to the global optimization of
  expensive-to-evaluate functions.
\newblock \emph{Journal of Global Optimization}, 43\penalty0 (2):\penalty0
  373--389, March 2009.

\bibitem[Wirsching et~al.(1995)Wirsching, Paez, and Ortiz]{RandomVibrations}
Paul~H. Wirsching, Thomas~L. Paez, and Keith Ortiz.
\newblock \emph{Random Vibrations, Theory and Practice}.
\newblock John Wiley \& Sons, Inc., 1995.

\end{thebibliography}

\end{document}